\pdfoutput=1

\documentclass[11pt]{article}

\usepackage[]{acl}

\usepackage{times}
\usepackage{latexsym}
\usepackage{multirow}
\usepackage{graphicx}
\usepackage{stfloats}

\usepackage[T1]{fontenc}

\usepackage[utf8]{inputenc}

\usepackage{microtype}

\title{\textit{SlovakBERT}: Slovak Masked Language Model}

\author{Matúš Pikuliak \and {Štefan Grivalský} \and \textbf{Martin Konôpka} \and \textbf{Miroslav Blšták} \\ \textbf{Martin Tamajka} \and \textbf{Viktor Bachratý} \and \textbf{Marián Šimko} \\
  Kempelen Institute of Intelligent Technologies \\
  \texttt{name.surname@kinit.sk} \\ \AND
  Pavol Balážik \and Michal Trnka \and
  \textbf{Filip Uhlárik} \\
  Gerulata Technologies \\
  \texttt{name.surname@gerulata.com} \\}

\begin{document}
\maketitle
\begin{abstract}
We introduce a new Slovak masked language model called \textit{SlovakBERT}. This is to our best knowledge the first paper discussing Slovak transformers-based language models. We evaluate our model on several NLP tasks and achieve state-of-the-art results. This evaluation is likewise the first attempt to establish a benchmark for Slovak language models. We publish the masked language model, as well as the fine-tuned models for part-of-speech tagging, sentiment analysis and semantic textual similarity.
\end{abstract}

\section{Introduction}

Fine-tuning pre-trained large-scale language models (LMs) is the dominant paradigm of current NLP. LMs proved to be a versatile technology that can help to improve performance for an array of NLP tasks, such as parsing, machine translation, text summarization, sentiment analysis, semantic similarity etc. The state-of-the-art performance makes LMs attractive for any language community that wants to develop its NLP capabilities. In this paper, we concern ourselves with the Slovak language and address the lack of language models, as well as the lack of established evaluation standards for this language.

In this paper, we introduce a new Slovak-only transformers-based language model called \textit{SlovakBERT}\footnote{Available at \url{https://github.com/gerulata/slovakbert}}. Although several multilingual models already support Slovak, we believe that developing Slovak-only models is still important, as it can lead to better results and more compute and memory-wise efficient processing of the Slovak language. \textit{SlovakBERT} has RoBERTa architecture~\cite{roberta} and it was trained with a Web-crawled corpus.

Since no standard evaluation benchmark for Slovak exists, we created our own set of tests mainly from pre-existing datasets. We believe that our evaluation methodology might serve as a standard benchmark for the Slovak language in the future. We evaluate \textit{SlovakBERT} with this benchmark and we also compare it to other available (mainly multilingual) LMs and other existing approaches. The tasks we use for evaluation are: part-of-speech tagging, semantic textual similarity, sentiment analysis and document classification. We also publish the best-performing models for selected tasks. These might be used by other Slovak researchers or NLP practitioners in the future as strong baselines.

Our main contributions in this paper are:

\begin{itemize}
    \setlength\itemsep{0em}
    \setlength\parskip{0em}
    \item We published a Slovak-only LM trained on a Web corpus.
    \item We established an evaluation methodology for the Slovak language and we apply it on our model, as well as on other LMs.
    \item We published several fine-tuned models based on our LM, namely a part-of-speech tagger, a sentiment analysis model and a sentence embedding model.
    \item We published several additional datasets for multiple tasks, namely sentiment analysis test sets and semantic similarity translated datasets (including a manually translated test set).
\end{itemize}

The rest of this paper is structured as follows: In Section~\ref{sec:related} we discuss related work about language models and their language mutations. In Section~\ref{sec:model} we describe the corpus crawling efforts and how we train \textit{SlovakBERT} with the resulting corpus. In Section~\ref{sec:evaluation} we evaluate the model with four NLP tasks.

\section{Related Work}\label{sec:related}

\subsection{Language Models}

LMs today are commonly based on self-attention layers called \textit{transformers}~\cite{transformer}. Despite the common architecture, the models might differ in the details of their implementation, as well as in the task they are trained with~\cite{whichbert}. Perhaps the most common task is the so-called \textit{masked language modeling}~\cite{bert}, where randomly selected parts of text are masked and the model is expected to fill these parts with the original tokens. Masked language models are useful mainly as backbones for further fine-tuning. Another approach is to train generative autoregressive models~\cite{gpt2}, which always predict the next word in a sequence, which can be used for various text generation tasks. Variants of LMs exist that attempt to make them more efficient~\cite{electra, tinybert}, able to handle longer sentences~\cite{longformer} or fulfill various other requirements.

\subsection{Availability in Different Languages}

English is the most commonly used language in NLP and a \textit{de facto} standard for experimental work. Most of the proposed LM variants are indeed trained and evaluated only on English. Other languages usually have at most only a few LMs trained, usually with a very safe choice of model architecture (e.g. BERT or RoBERTa). Languages with available native models are, to name only a few, French~\cite{french}, Dutch~\cite{dutch}, Greek~\cite{greek}, Arabic~\cite{arabic}, Czech~\cite{czech} or Polish~\cite{polish}.

There is no Slovak-specific large-scale LM available so far. There is a Slovak version of WikiBERT model~\cite{wikibert}, but it is trained only on texts from Wikipedia, which is not a large enough corpus for proper language modeling at this scale. The limitations of this model will be shown in the results as well.

\subsection{Multilingual Language Models}

Multilingual LMs are sometimes proposed as an alternative to training language-specific LMs. These LMs can handle more than one language, in practice often more than 100. Training them is more efficient than training separate models for all the languages. Additionally, cross-lingual transfer learning might improve the performance with the languages being able to learn from each other. This is especially beneficial for low-resource languages.

The first large-scale multilingual LM is MBERT~\cite{bert} trained on 104 languages. The authors observed that by simply exposing the model to data from multiple languages, the model was able to discover the multilingual signal and it spontaneously developed interesting cross-lingual capabilities, i.e. sentences from different languages with similar meanings also have similar representations. Other models explicitly use multilingual supervision, e.g. dictionaries, parallel corpora or machine translation systems~\cite{xlm, unicoder}. XLM-R~\cite{xlm-r} pushed the performance of multilingual LMs even further by increasing the scale of training by using Web-crawled data and a larger amount of compute.

\section{Training}\label{sec:model}

In this section, we describe our own Slovak masked language model -- \textit{SlovakBERT}, the data that were used for training, the architecture of the model and how it was trained.

\subsection{Data}

We used a combination of available corpora and our own Web-crawled corpus as our training data. The available corpora we used were: Wikipedia (326MB of text), Open Subtitles (415MB) and OSCAR 2019 corpus (4.6GB). We crawled \texttt{.sk} top-level domain webpages, applied language detection and extracted the title and the main content of each page as clean text without HTML tags (17.4GB). The text was then processed with the following steps:

\begin{itemize}
    \setlength\itemsep{0em}
    \setlength\parskip{0em}
    \item URL and email addresses were replaced with special tokens.
    \item Elongated punctuation was reduced, i.e. if there were sequences of the same punctuation mark, these were reduced to one mark (e.g. \texttt{-{}-} to \texttt{-}).
    \item Markdown syntax was deleted.
    \item All text content in braces \texttt{\{.\}} was eliminated to reduce the amount of markup and programming language text.
\end{itemize}

We segmented the resulting corpus into sentences and removed duplicates to get 181.6M unique sentences. In total, the final corpus has 19.35GB of text.

\subsection{Model Architecture and Training}

The model itself is a RoBERTa model~\cite{roberta}. The details of the architecture are shown in Table~\ref{tab:lms} in the \texttt{SlovakBERT} column. We use BPE~\cite{bpe} tokenizer with the vocabulary size of 50264. The model was trained for 300k training steps ($\approx$70 epochs) with a batch size of 512. Each epoch consists of approximately 4277 training steps. Samples were limited to a maximum of 512 tokens and for each sample, we fit as many full sentences as possible. We used Adam optimization algorithm~\cite{adam} with $5 \times 10^{-4}$ learning rate and 10k warmup steps. Dropout (dropout rate $0.1$) and weight decay ($\lambda = 0.01$) were used for regularization. We used \texttt{fairseq}~\cite{fairseq} library for training, which took approximately 248 hours on 4 NVIDIA A100 GPUs. We used 16-bit float precision.

\section{Evaluation}\label{sec:evaluation}

In this section, we describe the evaluation methodology and results for \textit{SlovakBERT} and other LMs. We conducted the evaluation on four different tasks: part-of-speech tagging, semantic textual similarity, sentiment analysis and document classification. For each task, we introduce the dataset that is used, various baseline solutions, the LM-based approach we took and the final results for the task. For some tasks (part-of-speech tagging and semantic textual similarity) we also performed layer-wise model analysis.

\subsection{Evaluated Language Models}

We evaluate and compare several LMs that support Slovak language to some extent:

\paragraph{\textbf{XLM-R}}~\cite{xlm-r} - XLM-R is a suite of multilingual RoBERTa-style LMs. The models support 100 languages, including Slovak. Training data are based on CommonCrawl Web-crawled corpus. The Slovak part has 23.2 GB (3.5B tokens). The XLM-R models differ in their size, ranging from Base model with 270M parameters to XXL model with 10.7B parameters.
\paragraph{\textbf{MBERT}}~\cite{mbert} - MBERT is a multilingual version of the original BERT model trained with Wikipedia-based corpus containing 104 languages. The authors do not mention the amount of data for each language, but considering the size of Slovak Wikipedia, we assume that the Slovak part has tens of millions of tokens.
\paragraph{\textbf{WikiBERT}}~\cite{wikibert} - WikiBERT is a series of monolingual BERT-style models trained on dumps of Wikipedia. The Slovak model was trained with 39M tokens. \\

Note that both XLM-R and MBERT models were trained in a cross-lingually unsupervised manner, i.e. no additional signal about how sentences or words from different languages relate to each other was provided. The models were trained with a multilingual corpora only, although language balancing was performed.

In Table~\ref{tab:lms} we provide basic quantitative measures for all the models. We compare their architecture and training data, and we also measure tokenization productivity (how many tokens are generated from given text) on \textit{Universal Dependencies}~\cite{ud} train set. We show the average length of tokens for each model. Longer tokens are considered to be better because they can be more semantically meaningful and also because they are more computationally efficient. We also show how many unique tokens were used (effective vocabulary) for the tokenization of this particular dataset. Multilingual LMs have a smaller portion of their vocabulary used since they contain many tokens useful mainly for other languages, but not for Slovak. These tokens are effectively redundant for Slovak text processing.

\begin{table*}[]
    \centering
    \small
    \begin{tabular}{l|l|l|l|l|l}
\textbf{Model} & SlovakBERT & XLM-R-Base & XLM-R-Large & MBERT & WikiBERT \\ \hline
Architecture & RoBERTa & \multicolumn{2}{c|}{RoBERTa} & BERT & BERT \\
Num. layers & 12 & 12 & 24 & 12 & 12 \\
Num. attention head & 12 & 12 & 16 & 12 & 12 \\
Hidden size & 768 & 768 & 1024 & 768 & 768 \\
Num. parameters & 125M & 278M & 560M & 178M & 102M \\
Languages & 1 & 100 & 100 & 104 & 1 \\
Training dataset size (tokens) & 4.6B & \multicolumn{2}{c|}{167B} & n/a & 39M \\
Slovak dataset size (tokens) & 4.6B & \multicolumn{2}{c|}{3.2B} & 25-50M & 39M \\
Vocabulary size & 50K & \multicolumn{2}{c|}{250K} & 120K & 20K \\
\\
\multicolumn{6}{l}{\textit{Universal Dependencies} train set tokenization} \\ \hline
Average token length (chars) & 3.23 & \multicolumn{2}{c|}{2.84} & 2.40 & 2.70 \\
Average word length (tokens) & 1.43 & \multicolumn{2}{c|}{1.63} & 1.93 & 1.71 \\
Effective vocabulary & 16.6K & \multicolumn{2}{c|}{9.6K} & 6.7K & 5.8K \\
Effective vocabulary (\%) & 33.05 & \multicolumn{2}{c|}{3.86} & 5.62 & 29.10 \\

    \end{tabular}
    \caption{Basic statistics about the evaluated LMs.}
    \label{tab:lms}
\end{table*}

\subsection{Part-of-Speech Tagging}

The goal of part-of-speech (POS) tagging is to assign a certain POS tag to each word. This task mainly evaluates the syntactic capabilities of the models.

\subsubsection{Data}

We use Slovak Dependency Treebank from \textit{Universal Dependencies} dataset~\cite{slovak-ud, ud} (UD). It contains annotations for both Universal (UPOS, 17 tags) and Slovak-specific (XPOS, 19 tags) POS tagsets. XPOS uses a more complicated system and it encodes not only POS tags, but also other morphological categories in the label. In this work, we only use the first letter from each XPOS label, which corresponds to a typical POS tag. The tagsets and their relations are shown in Table~\ref{tab:pos-tagsets}.

\subsubsection{Previous work}

Since Slovak is an official part of the UD dataset, systems that attempt to cover multiple or all UD languages often support Slovak as well. The following systems were trained on UD data and support both UPOS and XPOS tagsets:

\paragraph{\textbf{UDPipe 2}}~\cite{udpipe} - A deep learning model based on multilayer bidirectional LSTM architecture with pre-trained Slovak word embeddings. The model supports multiple languages, but the models themselves are monolingual.
\paragraph{\textbf{Stanza}}~\cite{stanza} - Stanza is a very similar model to UDPipe, it is also based on multilayer bidirectional LSTM with pre-trained word embeddings.
\paragraph{\textbf{Trankit}}~\cite{trankit} - Trankit is based on adapter-style fine-tuning~\cite{adapter} of XLM-R-Base. The adapters are fine-tuned for specific languages and they are able to handle multiple tasks at the same time. \\

\subsubsection{Our Fine-Tuning}

We use a standard setup for fine-tuning the LMs for token classification. The final layer of an LM that is used to predict the masked tokens is discarded. A classifier linear layer with dropout and softmax activation function is used in its place to generate a probability vector for each token. The loss function for the batch of samples is defined as an average cross-entropy across all the tokens. Note that there is a discrepancy between what we perceive as words and what the models use as tokens. Some words might be tokenized into multiple tokens. In that case, we only make the prediction on the first token, and the final classifier layer is not applied to the subsequent tokens for this word. We use \texttt{Hugging Face Transformers} library for LM fine-tuning.

\subsubsection{Results}

We have performed a random hyperparameter search with \textit{SlovakBERT}. The range of individual hyperparameters is shown in Table~\ref{tab:pos-hparams}. We have found out that weight decay is a beneficial regularization technique, while label smoothing proved itself to be inappropriate for our case. Other hyperparameters showed to have very little reliable effect, apart from the learning rate, which proved to be very sensitive. We have not repeated this tuning for other LMs, instead, we only tuned the learning rate. We have found out that it is appropriate to use a learning rate of $1 \times 10^{-5}$ for all the models, but XLM-R-Large. XLM-R-Large, the biggest model we tested, needs a smaller learning rate of $1 \times 10^{-6}$.

The results for POS tagging are shown in Table~\ref{tab:pos-results}. We report accuracy for both XPOS and UPOS tagsets. WikiBERT seems to be the worst-performing LM, probably because of its small training set. \textit{SlovakBERT} seems to be on par with larger XLM-R-Large. Other models lag behind slightly. From the existing solutions, only transformers-based Trankit seems to be able to keep up.

\begin{table}[]
    \small
    \centering
    \begin{tabular}{l|r|r}
 \textbf{Model} & \textbf{UPOS} & \textbf{XPOS} \\ \hline
 UDPipe 2.0 & 92.83 & 94.74 \\
 UDPipe 2.6 & 97.30 & 97.87 \\
 Stanza & 96.03 & 97.29 \\
 Trankit & 97.85 & 98.03 \\ \hline
 WikiBERT & 94.41 & 96.54 \\
 MBERT & 97.50 & 98.03 \\
 XLM-R-Base & 97.61 & 98.23 \\
 XLM-R-Large & \textbf{97.96} & 98.34 \\ \hline
 SlovakBERT & 97.84 & \textbf{98.37} \\
    \end{tabular}
    \caption{Results for POS tagging (accuracy).}
    \label{tab:pos-results}
\end{table}

We measured the POS performance for \textit{SlovakBERT} checkpoints (a checkpoint was made each 1000 steps) as well to see how soon the model acquired basic morphosyntactic capabilities.  We can see in Figure~\ref{fig:pos}, that the model was saturated w.r.t POS performance quite soon, after approximately 15k steps  ($\approx$3.5 epochs). We stopped the analysis after the first 125k steps   ($\approx$30 epochs) since the results seemed to be stable.

\subsubsection{Probing}

\begin{figure}[t]
\centering
\includegraphics[width=7.8cm]{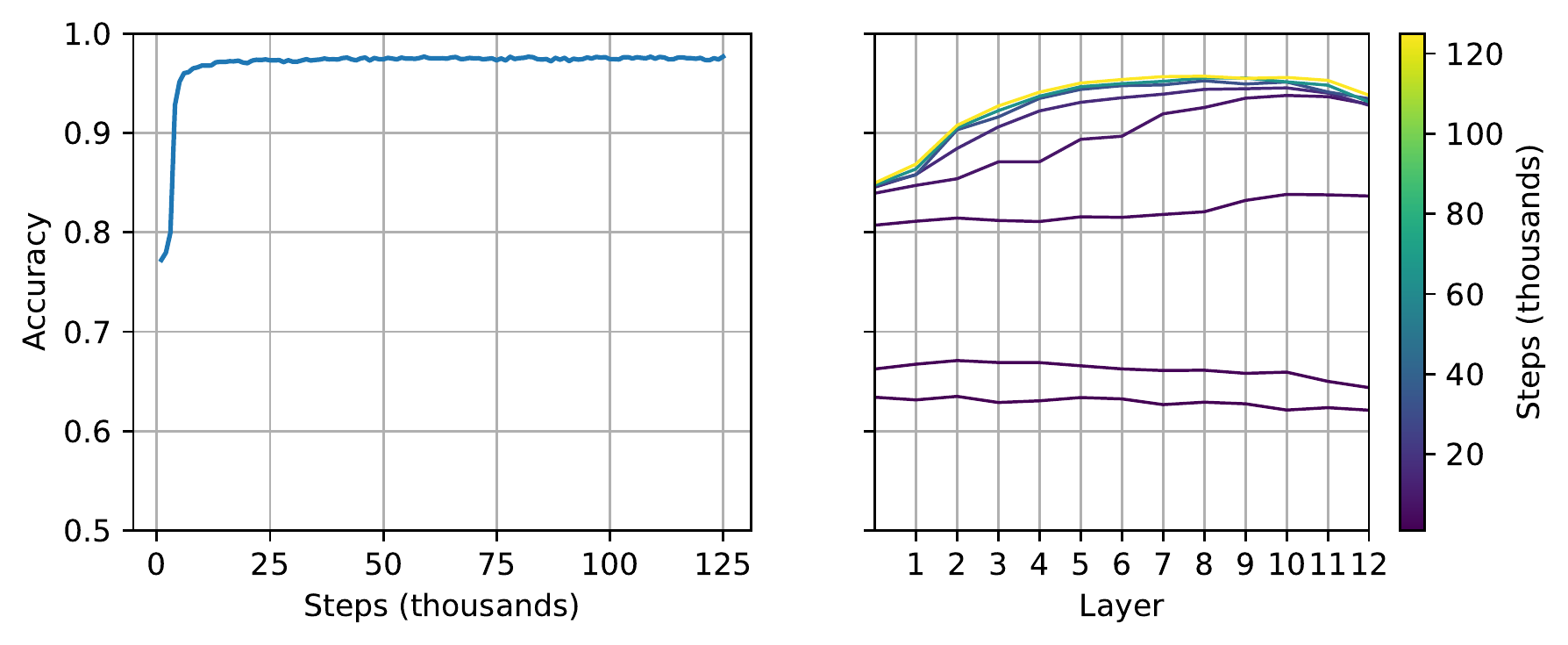}
\caption{Analysis of POS tagging learning dynamics. \textit{Left:} Accuracy after fine-tuning the different checkpoints. \textit{Right:} Accuracy of probes on all the layers of different checkpoints. Each line represents one checkpoint and its results on all the layers.}\label{fig:pos}
\end{figure}

We performed probing by training linear classifier on representations from individual layers of frozen models~\cite{belinkov-etal-2017-neural}. We show the performance of these probes for all the layers for checkpoints from \textit{SlovakBERT} training in Figure~\ref{fig:pos}. The probing is done on models that are \textit{not} fine-tuned for POS tagging. Layer-wise, the performance peaks quite soon at layer 6 and then plateaus. The last layers even have degraded performance. The results are in accord with the current understanding of how LMs work, i.e. that they process text in a bottom-up manner and the morphosyntactic information needed for POS tagging is being processed mainly in the middle layers~\cite{bert-pipeline}. We can also see that the performance for individual layers peaks quite soon during the training, with a slight lag for earlier layers.

\subsection{Semantic Textual Similarity}

Semantic textual similarity (STS) is an NLP task where similarity between pairs of sentences is measured. In our work, we train the LMs to generate sentence embeddings and then we measure how much the cosine similarity between embeddings correlates with the ground truth labels provided by human annotators. We can use the resulting models to generate universal sentence embeddings for Slovak.

\subsubsection{Data}

Currently, there is no native Slovak STS dataset. We decided to machine translate existing English datasets STSbenchmark~\cite{stsb} and SICK~\cite{sick} into Slovak. These datasets use a $\langle 0,5 \rangle$ scale that expresses the similarity of two sentences. The meaning of individual steps on this scale is shown in Table~\ref{tab:sts-labels}. We used M2M100 (1.2B parameters variant) machine translation system~\cite{m2m100}. The test set from STSbenchmark was manually translated by the authors. These translations were used for evaluation only and are published as well.

\subsubsection{Previous Work}

No Slovak-specific sentence embedding model has been published yet. We use a naive solution based on Slovak word embeddings and several available multilingual models for comparison:

\paragraph{\textbf{fastText}}~\cite{fasttext} - We use pre-trained Slovak fastText word embeddings to generate representations for individual words. The sentence representation is an average of all its words. This represents a very naive baseline since it completely omits the word order.
\paragraph{\textbf{LASER}}~\cite{laser} - LASER is a model trained to generate multilingual sentence embeddings. It is based on an encoder-decoder LSTM machine translation system that is trained with 93 languages. The encoder is shared across all the languages and as such, it is able to generate multilingual representations.
\paragraph{\textbf{LaBSE}}~\cite{labse} - LaBSE is an MBERT model fine-tuned with parallel corpus to produce multilingual sentence representations.
\paragraph{\textbf{XLM-R$_{EN}$}}~\cite{multi-sbert} - XLM-R model fine-tuned with English STS-related data (SNLI, MNLI and STSbenchmark datasets). This is a zero-shot cross-lingual learning setup, i.e. no Slovak data are used and only English fine-tuning is done.

\subsubsection{Our Fine-Tuning}

We use a setup similar to~\cite{multi-sbert}. A pre-trained LM is used to initialize a Siamese network. Both branches of the network are identical LMs with a mean-pooling layer at the top that generates the final sentence embeddings. The embeddings from the two sentences are compared using cosine similarity. The network is trained as a regression model, i.e. the final computed similarity is compared with the ground truth similarity with \textit{mean squared error} loss function. We use \texttt{SentenceTransformers} library for the fine-tuning.

\subsubsection{Results}

We compare the systems using Spearman correlation between the cosine similarity of the generated sentence representations and the ground truth data. The original STS datasets are using $\langle 0,5 \rangle$ scale. We normalize these scores to $\langle 0,1 \rangle$ range so that they can be directly compared to the cosine similarities. We performed a hyperparameter search in this case as well. Again, we have found out that the results are quite stable across various hyperparameter values, with the learning rate being the most sensitive hyperparameter. The details of the hyperparameter tuning are shown in Table~\ref{tab:sts-hparams}. We show the main results in Table~\ref{tab:sts-results}.

\begin{table}[]
    \small
    \centering
    \begin{tabular}{l|r|r}
          & \multicolumn{2}{c}{\textbf{Translation}} \\
         \textbf{Model} & \textbf{Manual} & 
          \textbf{M2M100}\\ \hline
         fastText & 0.366 & 0.383 \\
         LASER & 0.706 & 0.711 \\
         LaBSE & 0.730 & 0.739 \\
         XLM-R$_{EN}$ & \textbf{0.804} & \textbf{0.801} \\ \hline
         WikiBERT & 0.652 & 0.673 \\
         MBERT & 0.726 & 0.734 \\
         XLM-R-Base & 0.785 & 0.791 \\
         XLM-R-Large & 0.794 & 0.790 \\ \hline
         SlovakBERT & 0.793 & 0.799 \\
    \end{tabular}
    \caption{Spearman correlation between cosine similarity of generated representations and desired similarities on STSbenchmark dataset translated to Slovak.}
    \label{tab:sts-results}
\end{table}

We can see that the results are fairly similar to POS tagging w.r.t. how the LMs are relatively ordered. The existing solutions are worse, except for XLM-R$_{EN}$ trained with English data, which is actually the best-performing model in our experiments. It seems that their model fine-tuned with real data without machine-translation-induced noise works better, even if it has to perform the inference cross-lingually on Slovak data. We have found out that manual translation of the test set did not yield significantly different results compared to machine translation, despite the fact that most of the machine-translated samples were quite noisy according to our manual inspection. This shows that we can measure how good STS systems are even with noisy machine-translated data.

We also experimented with Slovak-translated NLI data in a way where the model was first fine-tuned on the NLI task and then the final STS fine-tuning was performed. However, we were not able to outperform the purely STS fine-tuning with this approach and the results remained virtually the same. This result is in contrast with the usual case for English training, where the NLI data regularly improve the results~\cite{sbert}. We theorize that this effect might be caused by noisy machine translation.

\begin{figure}[t]
\centering
\includegraphics[width=7.8cm]{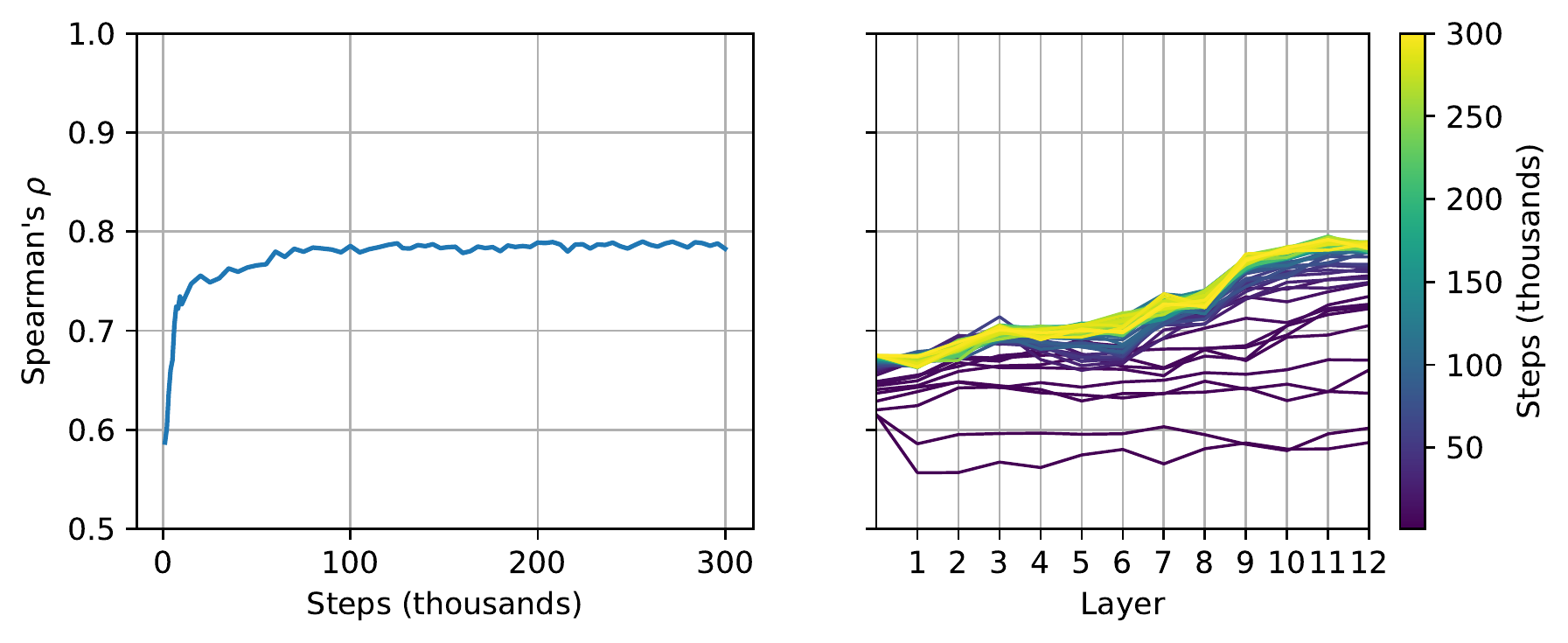}
\caption{Analysis of STS learning dynamics. \textit{Left:} Spearman correlation after fine-tuning with various checkpoints. \textit{Right:} Spearman correlation on all the layers with selected checkpoints. Each line represents one checkpoint and its results on all the layers.}\label{fig:sts}
\end{figure}

Figure~\ref{fig:sts} shows the learning dynamics of STS. On the left, we can see that the performance takes much longer to plateau than in the case of POS. This shows that the model needs longer time to learn about semantics. Still, we can see that the performance ultimately stabilizes just below $0.8$ score.

We also performed a layer-wise analysis, where we analyzed which layers have the most viable representations for this task. We conducted the mean-pooling at different layers and ignored all the subsequent layers. We can see that the best-performing layers are actually the last layers of the model.

\subsection{Sentiment Analysis}\label{sec:sa}

The goal of sentiment analysis is to identify the affective sentiment of a given text. It requires semantic analysis of the text, as well as a certain amount of emotional understanding.

\subsubsection{Data}

We use a Twitter-based dataset~\cite{twitter-sa} annotated on a scale with three values: \textit{negative}, \textit{neutral} and \textit{positive}. Some of the tweets have already been removed since the dataset was created. Therefore, we work with a subset of the original dataset.

We cleaned the data by removing URLs, retweet prefixes, hashtags, user mentions, quotes, asterisks, redundant whitespaces and trailing punctuation. We have also deduplicated the samples, as there were cases of identical samples (i.e. retweets) or very similar samples (i.e. automatically generated tweets). These duplicates had in some cases different labels. After the deduplication, we were left with 41084 tweets with 11160 negative samples, 6668 neutral samples and 23256 positive samples.

Additionally, we have also manually annotated a series of test sets containing reviews from various domains: accommodation, books, cars, games, mobiles and movies. Each domain has approximately 100 manually labeled samples. These are published along with this paper. They serve to check how well the model behavior transfers to other domains. This dataset is called \textit{Reviews} in the results below, while the original Twitter-based dataset is called \textit{Twitter}.

\subsubsection{Previous Work and Baselines}

The original paper introducing the Twitter dataset introduced an array of traditional classifiers (Naive Bayes and 5 SVM variants) to solve the task. The authors report macro-F1 score for positive and negative classes only. Additionally, unlike us, they worked with the whole dataset. Approximately 10K tweets have been deleted since the dataset was introduced. \cite{pecar} use the same version of the dataset as we do. They use approaches based on word embeddings and ELMO~\cite{elmo} to solve the task. Note that both published works use cross-validation, but no canonical dataset split is provided in either of them.

There are several existing approaches we use for comparison:

\paragraph{\textbf{NLP4SK}\footnote{\url{http://arl6.library.sk/nlp4sk/webapi/analyza-sentimentu}}} - A rule-based sentiment analysis system for Slovak that is available online
\paragraph{\textbf{Amazon}} - We also translated the Slovak data into English and used Amazon's commercial sentiment analysis API and tested its performance on our test sets. \\

We implemented several baseline classifiers that were trained with the same training data as the LMs in our experiments:

\paragraph{\textbf{TF-IDF linear classifier}} - A perceptron trained with SGD algorithm. The text is represented with TF-IDF using N-grams as basic text units.
\paragraph{\textbf{fastText classifier}} - We used the built-in fastText classifier with and without pre-trained Slovak word embedding models.
\paragraph{\textbf{Our STS embedding linear classifier}} - A perceptron trained with SGD algorithm. The text is represented using the sentence embedding model we have trained for STS. \\

We performed a random search hyperparameter optimization for all the approaches.

\subsubsection{Our Fine-Tuning}

We fine-tuned the LMs as classifiers with 3 classes. The topmost layer of an LM is discarded and instead, a multilayer perceptron classifier with one hidden layer and dropout is applied to the representation of the first token. The categorical cross-entropy loss function is used as the loss function. The class with the highest probability coming from the softmax function is selected as the predicted label during inference. We use \texttt{Hugging Face Transformers} library for fine-tuning.

\subsubsection{Results}

We report macro-F1 scores for all three classes as our main performance measure. The LMs were trained on the Twitter dataset. We calculate the average F1 from our \textit{Reviews} dataset as an additional measure.

Again, we have performed a hyperparameter optimization of \textit{SlovakBERT}. The results are similar to results from POS tagging and STS. We have found out that the learning rate is the most sensitive hyperparameter and that a small amount of weight decay is a beneficial regularization. The main results are shown in Table~\ref{tab:sa-results}. We can see that we were able to obtain better results than the results that had been reported previously. However, the comparison is not perfect, as we use slightly different datasets for the aforementioned reasons.

\begin{table}[]
    \small
    \centering
    \begin{tabular}{l|r|r|r}
        \textbf{Model} & \multicolumn{2}{c|}{\textbf{Twitter F1}} & \textbf{Reviews F1} \\
        & \textbf{3-class} & \textbf{2-class} & \textbf{3-class}\\ \hline
        \cite{twitter-sa}* & - & 0.682 & -  \\
         \cite{pecar}* & 0.669 & - & - \\ 
         Amazon & 0.502 & 0.472 & 0.766 \\
         NLP4SK & 0.489 & 0.468 & \textbf{0.815} \\ \hline
         TF-IDF & 0.571 & 0.603 & 0.412 \\
         fastText & 0.591 &  0.622 &  0.416 \\
         fastText w/ emb. & 0.606 &  0.631 &  0.426 \\
         STS embeddings & 0.581 & 0.597 & 0.582 \\ \hline
         WikiBERT & 0.580 & 0.597 & 0.398 \\
         MBERT & 0.587 & 0.622 & 0.453 \\
         XLM-R-Base & 0.620 & 0.651 & 0.518 \\
         XLM-R-Large & 0.655 & \textbf{0.716} & 0.617 \\ \hline
         SlovakBERT & \textbf{0.672} & 0.705 & 0.583 \\
    \end{tabular}
    \caption{Macro-F1 scores for sentiment analysis task. The 2-class F1 score for Twitter is calculated only from positive and negative classes -- a methodology introduced in the original dataset paper. *Indicates different evaluation sets.}
    \label{tab:sa-results}
\end{table}

The LMs are ordered in performance similarly to how they are ordered in the two previous tasks. \textit{SlovakBERT} seems to be among the best performing models, along with the larger XLM-R-Large. The LMs were also able to successfully transfer their sentiment knowledge to new domains and they achieve up to 0.617 macro-F1 in the reviews as well. However, both Amazon commercial sentiment API and NLP4SK have even better scores, even though their performance on Twitter data was not very impressive. This is probably caused by the underlying training data they use in their systems, which might match our \textit{Reviews} datasets more than the tweets used for our fine-tuning.

\subsection{Document Classification}

The final task which we evaluate our LMs on is a classification of documents into 5 news categories. The goal of this task is to ascertain how well LMs handle common classification problems. We use a Slovak Categorized News Corpus~\cite{scnc} that contains 4.7K news articles classified into 6 classes: Sports, Politics, Culture, Economy, Health and World. We do not use the \textit{Culture} category, since it contains a significantly smaller number of samples.

Unfortunately, no existing work has used this dataset for document classification, so there are no existing results publicly available. We use the same set of baselines and LM fine-tuning as in the case of sentiment analysis since both these tasks are text classification tasks, see Section~\ref{sec:sa} for more details.

\subsubsection{Results}

The main results from our experiment are shown in Table~\ref{tab:dc-results}. We can see that the LMs are again the best-performing approach. In this case, the results are quite similar with \textit{SlovakBERT} being the best by a narrow margin. The baselines achieved significantly worse results. Note that our sentence embedding model has the worst results on this task, while it had competitive performance in sentiment classification. We theorize, that the sentence embedding model was trained on sentences and is, therefore, less capable of handling longer texts, typical for the dataset used here.

\begin{table}[]
    \small
    \centering
    \begin{tabular}{l|r}
        \textbf{Model} & \textbf{F1} \\
         TF-IDF & 0.953 \\
         fastText & 0.963 \\
         fastText w/ emb. & 0.963 \\
         STS embeddings & 0.935 \\ \hline
         WikiBERT & 0.935 \\
         MBERT & 0.985 \\
         XLM-R-Base & 0.987 \\
         XLM-R-Large & 0.985 \\ \hline
         Our model & \textbf{0.990} \\
    \end{tabular}
    \caption{Macro-F1 scores for document classification task.}
    \label{tab:dc-results}
\end{table}

\section{Conclusions}\label{sec:concl}

We have trained and published \textit{SlovakBERT} -- a new large-scale transformers-based Slovak masked language model using 19.35GB of Web-crawled Slovak text. We proposed an evaluation benchmark with multiple tasks for Slovak language and evaluated several models. We conclude, that \textit{SlovakBERT} achieves state-of-the-art results on this benchmark, but multilingual language models are still competitive, especially larger but computationally less efficient models such as XLM-R-Large. We also release fine-tuned models for the Slovak community.

The lack of evaluation benchmarks is still an issue for many mid-resource languages, i.e. languages that have a sizeable corpus of text available on the Web, but do not have annotated natural language understanding datasets available. Our work was limited by this as well, as we were forced to use datasets created by machine translation (in case of STS), noisy datasets (in case of sentiment analysis), or datasets with almost saturated performance (in case of document classification). Creating new high-quality datasets for the evaluation of Slovak is our future work.

\section{Limitations}

\paragraph{Limited performance evaluation.} As we have already noted in Section 5, the lack of annotated data limits our ability to properly evaluate the performance of \textit{SlovakBERT}. The existing datasets have some issues:

\begin{itemize}
    \item \textit{UD for POS tagging.} The performance for POS tagging seems already saturated.
    \item \textit{Sentiment analysis.} The training data for sentiment analysis are quite noisy and many samples are misclassified. The data are also based on Twitter and as such, they are not easily accessible and the number of samples available is constantly decreasing. This is a threat to the replicability of our results.
    \item \textit{Semantic textual similarity.} We use translated English STS data for training. The translation itself is a noisy process. The data are also US-centric in their nature and they might not exactly match the needs of Slovak speakers.
    \item \textit{Document classification.} The performance for this task is saturated as well.
\end{itemize}

We addressed some issues by manually creating evaluation sets for both sentiment analysis and semantic textual similarity. In the future, it would be appropriate to develop new datasets for higher-level NLP tasks, such as natural language inference or question answering.

\paragraph{Limited ethical evaluation.} For similar reasons as above, there is no evaluation of bias in the Slovak-processing language models. As of now, it is not clear how biased the models are, since evaluation benchmarks were not yet designed for the Slovak language. We made note of this issue in the \textit{Ethical Consideration} section as well.

\section{Ethical Consideration}

\textit{SlovakBERT} was trained using a Web-crawled corpus. This is a common practice in current NLP, yet, it raises some ethical concerns. Models trained with huge poorly documented corpora might encode in them various societal biases. The Slovak texts written on the Web are not representative of all Slovak users. Certain demographic groups might be underrepresented and the model might not reflect them accordingly. We do not study these effects in this work and we do not recommend using our model for sensitive applications without further analysis. Unfortunately, there are no datasets, benchmarks, or other resources able to measure these effects in the Slovak language as of yet.

\section*{Acknowledgments}
This research was partially supported by the Central European Digital Media Observatory (CEDMO), a project funded by the European Union under the Contract No. 2020-EU-IA-0267.

\bibliography{acl_latex}
\bibliographystyle{acl_natbib}

\appendix
\clearpage
\section{Hyperparameter Values}

\begin{table}[h]
    \centering
    \small
    \begin{tabular}{l|r|r}
  \textbf{Hyperparameter} & \textbf{Range} & \textbf{Selected} \\ \hline
  Learning rate & $[10^{-7}, 10^{-3}]$ & $10^{-5}$ \\
  Batch size & $\{8, 16, 32, 64, 128\}$ & $32$ \\
  Warmup steps & $\{0, 500, 1000, 2000\}$ & $1000$ \\
  Weight decay & $[0, 0.1]$ & $0.05$ \\
  Label smoothing & $[0, 0.2]$ & $0$ \\
  Learning rate scheduler & Various\footnotemark & \texttt{linear} \\
    \end{tabular}
    \caption{Hyperparameters used for POS tagging. Adam was used as an optimization algorithm.}
    \label{tab:pos-hparams}
\end{table}
\footnotetext{See the list of schedulers supported by Hugging Face Transformers library.}

\begin{table}[h]
    \centering
    \small
    \begin{tabular}{l|r|r}
  \textbf{Hyperparameter} & \textbf{Range} & \textbf{Selected} \\ \hline
  Learning rate & $[10^{-7}, 10^{-3}]$ & $10^{-5}$ \\
  Batch size & $\{8, 16, 32, 64, 128\}$ & $32$ \\
  Warmup steps & $\{0, 500, 1000, 2000\}$ & $1000$ \\
  Weight decay & $[0, 0.2]$ & $0.15$ \\
  Learning rate scheduler & Various\footnotemark & \texttt{cosine with hard restarts} \\
    \end{tabular}
    \caption{Hyperparameters used for STS tagging. Adam was used as an optimization algorithm.}
    \label{tab:sts-hparams}
\end{table}
\footnotetext{See the list of schedulers supported by the Sentence Transformers library.}

\clearpage
\section{Tagging Schemata}

\begin{table}[h]
    \centering
    \small
    \begin{tabular}{l|l||l|l}
\multicolumn{2}{c||}{\textbf{XPOS}} & \multicolumn{2}{c}{\textbf{UPOS}}  \\ \hline 
 \textbf{Tag} & \textbf{Description} & \textbf{Tag} & \textbf{Description} \\ \hline \hline
 A & adjective & \multirow{2}*{ADJ} & \multirow{2}*{adjective} \\ \cline{1-2}
 G & participle & & \\ \hline
 E & preposition & ADP & adposition \\ \hline
 D & adverb & ADV & adverb \\ \hline
 Y & conditional morpheme & \multirow{2}*{AUX} & \multirow{2}*{auxiliary} \\ \cline{1-2} 
 \multirow{2}*{V} & \multirow{2}*{verb} & & \\ \cline{3-4} 
 & & VERB & verb \\ \hline
 \multirow{2}*{O} & \multirow{2}*{conjuction} & CCONJ & coordinating conjunction \\ \cline{3-4} 
 & & SCONJ & subordinating conjunction \\ \hline
  \multirow{2}*{P} & \multirow{2}*{pronoun} & DET & determiner \\ \cline{3-4}
 & & \multirow{2}*{PRON} & \multirow{2}*{pronoun} \\ \cline{1-2}
 R & reflexive pronoun & & \\ \hline
 J & interjection & INTJ & interjection \\ \hline
  \multirow{2}*{S} & \multirow{2}*{noun} & NOUN & noun \\ \cline{3-4}
 & & PROPN & proper noun \\ \hline
 N & numeral & \multirow{2}*{NUM} & \multirow{2}*{numeral} \\ \cline{1-2}
 0 & digit & & \\ \hline
 T & particle & PART & particle \\ \hline
 Z & punctuation & PUNCT & punctuation \\ \hline
 W & abbreviation & \multirow{4}*{X} & \multirow{4}*{other} \\ \cline{1-2}
 Q & unidentifiable & & \\ \cline{1-2}
 \# & non-word element & & \\ \cline{1-2}
 \% & citation in foreign language & & \\ \hline
 & & SYM & symbol \\
    \end{tabular}
    \caption{Slovak POS tagsets and their mapping~\cite{slovak-ud}.}
    \label{tab:pos-tagsets}
\end{table}

\begin{table}[h]
    \centering
    \small
    \begin{tabular}{l|p{10cm}}
         \textbf{Label} & \textbf{Meaning} \\ \hline
         0 & The two sentences are completely dissimilar.  \\ \hline
         1 & The two sentences are not equivalent, but are on the same topic. \\ \hline
         2 & The two sentences are not equivalent, but share some details. \\ \hline
         3 & The two sentences are roughly equivalent, but some important information differs.\\ \hline
         4 & The two sentences are mostly equivalent, but some unimportant details differ. \\ \hline
         5 & The two sentences are completely equivalent, as they mean the same thing. \\
    \end{tabular}
    \caption{Annotation schema for STS datasets~\cite{sick}.}
    \label{tab:sts-labels}
\end{table}

\end{document}